# Automatic Real-word Error Correction in Persian Text

Seyed Mohammad Sadegh Dashti[1], Amid Khatibi Bardsiri[1], Mehdi Jafari Shahbazzadeh[2]

[1] Computer Engineering Department, Kerman Branch, Islamic Azad University,
Kerman, Iran

[2] Electrical Engineering Department, Kerman Branch, Islamic Azad University,
Kerman, Iran.

Corresponding Author: Amid Khatibi Bardsiri; a.khatibi@srbiau.ac.ir

**Abstract**

Automatic spelling correction stands as a pivotal challenge within the ambit of natural language processing (NLP), demanding nuanced solutions. Traditional spelling correction techniques are typically only capable of detecting and correcting non-word errors, such as typos and misspellings. However, context-sensitive errors, also known as real-word errors, are more challenging to detect because they are valid words that are used incorrectly in a given context. The Persian language, characterized by its rich morphology and complex syntax, presents formidable challenges to automatic spelling correction systems. Furthermore, the limited availability of Persian language resources makes it difficult to train effective spelling correction models. This paper introduces a cutting-edge approach for precise and efficient real-word error correction in Persian text. Our methodology adopts a structured, multi-tiered approach, employing semantic analysis, feature selection, and advanced classifiers to enhance error detection and correction efficacy. The innovative architecture discovers and stores semantic similarities between words and phrases in Persian text. The classifiers accurately identify real-word errors, while the semantic ranking algorithm determines the most probable corrections for real-word errors, taking into account specific spelling correction and context properties such as context, semantic similarity, and edit-distance measures. Evaluations have demonstrated that our proposed method surpasses previous Persian real-word error correction models. Our method achieves an impressive F-measure of 96.6% in the detection phase and an accuracy of 99.1% in the correction phase. These results clearly indicate that our approach is a highly promising solution for automatic real-word error correction in Persian text.

**Keywords:** real-word error, spelling correction, language modelling, semantic similarity, Persian language

## 1. Introduction

Spelling correction software is an essential component of text processing environments, commonly used to detect orthographically incorrect words, known as non-word errors. Traditional spelling correction software helps users replace erroneous words by suggesting appropriate correction candidates from a dictionary list. These systems verify user input words against the dictionary, treating words found in the lexicon as correct, irrespective of their context within the sentence. Although this method can identify misspellings, its effectiveness diminishes when detecting context-sensitive errors or 'real-word' errors.

Real-word errors arise when users inadvertently type a lexically correct word, albeit not the one intended, confirmed as valid by the dictionary. These errors can change the meaning of sentences and are commonly the result of typographical errors, such as insertion, deletion, substitution, and transposition of characters. Word boundary problems, including agglutination and split errors, further complicate error correction. Real-word errors can also happen when a user selects a correction candidate from a suggested word list generated by word processing applications and spelling correction software [1]. Additionally, spelling correction software featuring auto-correction options may replace non-word errors with improper replacement candidates, leading to real-word errors [2]. Other systems providing textual output, such as ASR (Automatic Speech Recognition) and OCR (Optical Character Recognition) software, are also prone to errors when used in inadequate environments. ASR systems are prone to generating orthographic and context-sensitive errors due to excessive noise in the environment, the speech quality, the accent, the context, and the vocabulary size of the system. [3, 4]. OCR systems are also erroneous and generate both non-word and real-word errors in the output text, especially when the scanned images are of low quality [5, 6]. In English text, real-word errors account for about 25% to 40% of all spelling errors [7, 8]. Real-word errors pose a significant challenge in the correction of Persian text due to the language's extensive vocabulary and intricate properties. One of the primary obstacles in real-word error correction for Persian text is the scarcity of available corpora for context errors. To mitigate this issue, researchers have introduced artificially generated real-word errors into regular texts. Moreover, the Persian language showcases distinct characteristics that exacerbate the complexity of real-word error correction, such as homophony, polysemy, and heterography. Homophony pertains to words that are phonetically identical but semantically distinct, while polysemy concerns words with multiple meanings. Heterography involves words with identical spellings but different meanings contingent on their pronunciation. Word boundary issues and pseudo-space further complicate Persian language processing, hindering the accurate detection and correction of real-word errors in Persian text. Despite these challenges, efforts have been made to develop statistical and rule-based methods for detecting and correcting real-word errors in Persian, with limited success. Real-word error correction in Persian remains an active area of research, and the development of more advanced methods and tools may help overcome these challenges and improve accuracy. This study aims to present a state-of-the-art approach for real-word error detection and correction in Persian. we outline our key contributions as follows:

- Semantic Architecture: Our developed semantic architecture introduces a sophisticated methodology for constructing and preserving similarity graphs tailored to word sequences from varied corpora. Its notable strength lies in its capacity to facilitate nuanced semantic analysis, ensuring robustness and extensibility in processing linguistic data.

- Error Detection: Classifiers are trained with features mainly from semantic architecture to identify real-word errors in Persian text. The classifiers are trained

- on a corpus of Persian text that has been annotated with real-word errors.

- Error Correction: The most likely corrections for real-word errors are ranked using a semantic ranking algorithm. The semantic ranker takes into account the context of the error and the meaning of the text.

- Evaluation Measures: We assess the effectiveness of our hybrid approach by using evaluation metrics such as F1 score. We also compare our approach to existing techniques for real-word error detection and correction.

Rest of this article is organized as follows: first, we present a review of previous work in the field. Next, we elaborate on challenges of Persian language and its properties. We then explain the proposed approach. After that, evaluation and experimental results are provided and discussed. Finally, in the concluding section, we synthesize our key findings and suggest avenues for future research, thereby encapsulating the comprehensive insights derived from our analysis.

## 2. Related Works

Automatic correction of word errors is a key task in NLP systems and text-based environments. Since the early 1960s, many approaches have been developed to tackle this problem. Edit distance-based models, such as the Damerau [9] and Damerau-Levenshtein method [10], used to be popular techniques to deal with errors in isolation. Phonetic algorithms such as Aspell[1] [11] and Jazzy[2] [12], word frequency approaches, [13-15] and noisy channel models, [16-20], have also been developed to improve accuracy for isolated word error corrections. Work in the field of automatic correction of word errors is not limited to isolated error corrections. Using context information has proven to be effective in increasing the performance of auto-correction of errors. For example, one approach used weighted combinations of conceptual density and N-grams to auto-correct misspellings in free running English text [21]. However, detection and correction of real-word errors is more complicated. Different assumptions such as predefined confusion sets, limited numbers of corrections, and other constraints are used to tackle the problem.

Context measures such as semantic distance [2] and noisy channel models based on N-grams have been employed in many NLP applications. Authors in [22] proposed a method for detecting not safe for work (NSFW) content on Reddit using a combination of natural language processing and machine learning techniques. Recently, a multi-dimensional definition of scope and a general framework for measuring the sentiment scope of a user on any social network were introduced [23]. These measures have also been used to detect and correct real-word errors [1, 24-26]. One of the early approaches in the 1990s leveraged context information by means of N-grams [24]. Also a new variation of Mays and colleagues' method was developed using fixed window sizes to achieve better performance [1]. In another work, local word bigrams and trigrams were used to estimate the probability score of candidate words [25]. A new approach was also developed to correct multiple real-word errors in situations where the source of the error was too noisy [27]. This approach used a combination of constraint grammars and shifting windows to detect multiple errors in a single sentence. Dashti developed a model that addressed the detection and auto-correction of real-word errors in cases where more than one error rested in a given word sequence [28]. The model leveraged probabilistic context-free grammars to distinguish syntactically well-formed word sequences in the search space. Word boundary challenges for both non-word and real-word errors are not frequently addressed. Flor categorized four different types of error corrections, including single-token and multi-token corrections for both non-word and real-word errors in student essays [17]. Kilicoglu and colleagues used split and merge to resolve aggregation and split errors [29].

---

[1] available at: http://aspell.net

[2] available at: https://sourceforge.net/projects/jazzy

State-of-the-art approaches leverage information from context using neural word or sense embeddings for spelling correction. A method based on search space reduction has also been proposed for the task of English language spelling correction [30]. This method has shown promising performance, both in terms of search space reduction and error correction. The use of spelling correction techniques is not limited to free running English text; they have also been applied to other domains, such as medical text.

Kilicoglu et al. have developed a model that combines edit distance similarity scores, frequency counts, and contextual similarity obtained from word embeddings for detecting and correcting different types of errors [29]. An unsupervised context-sensitive approach for clinical spelling correction has also been proposed [31]. This method uses word and character N-gram embeddings and leverages a ranking algorithm that can be tuned to match a specific domain. A dual embedding model has been developed to automatically correct misspellings in consumer health inquiries[32]. In this work, a pipeline was proposed where different sets of approaches were leveraged at each stage to handle various types of errors.

Pretrained contextual embeddings have also been used to detect and correct real-word errors [33]. In 2020, a study was published that aimed to solve the context-sensitive spelling error problem for English documents [34]. The study focused on typographical errors and used a deep learning method to solve the context-sensitive spelling error problem. The deep learning language model-based correction approach was divided into four parts: correction based on word embedding information, contextual embedding information, an auto-regressive (AR) language model, and an auto-encoding (AE) language model. In another work, a BERT-Based model was developed [35]. The proposed model first used BERT to represent the context of the misspelled word. Then, it used a soft-masked attention mechanism to attend to the representations of the correct words in the dictionary. The soft-masked attention mechanism allowed the model to focus on the most relevant words in the dictionary, which helped improve the accuracy of the correction. NeuSpell is a neural spelling correction toolkit that provides a variety of pre-trained models and evaluation tools [36]. It was designed to be easy to use and adaptable to different tasks and datasets. NeuSpell has been used for a variety of spelling correction tasks, including correcting typos, homophones, and grammatical errors. SpellBERT is a lightweight pre-trained model for Chinese spelling check, based on BERT and fine-tuned on a dataset of misspelled words [37]. It has been shown to be effective for Chinese spelling check and more efficient than other large language models. A recent work introduced a pre-training approach for Chinese spelling correction that used misspelled knowledge. The approach first created a confusion set of misspelled words, then pre-trained a Transformer model on a large corpus of text and the confusion set. The proposed model improved the performance of spelling correction for Chinese and outperformed other baseline methods [38]. Authors in [39] proposed a phonetic pre-training approach for Chinese spelling correction. The approach first pre-trained a Transformer model on a large corpus of text and phonetic information, then fine-tuned the model on a dataset of misspelled words. Tran and colleagues proposed a new context-sensitive spelling correction model for clinical text, using a conditional independence method to learn the relationships between words and their misspellings in clinical text [40]. A contextual spelling correction approach for customization of end-to-end speech recognition systems was also introduced, using a Transformer model to learn the relationships between words and their misspellings in the context of a speech recognition system [41]. This model improved the performance of spelling correction for end-to-end speech recognition systems. A multi-task detector-corrector framework for Chinese spelling correction was proposed in [42]. The

MDCSpell framework used a detector to identify misspelled words and a corrector to correct them. Liu et al. proposed a contextual typo robust approach, CRASpell, to improve Chinese spelling correction [43]. It used a context encoder to extract the context of a misspelled word and a corrector to correct it. An error-guided correction model for Chinese spelling error correction was introduced in [35]. It identified errors in text, generated a list of possible corrections for each error using a neural network, and selected the most likely correction given the surrounding text. This approach was effective for Chinese spelling error correction and outperformed other baseline methods. AraSpell, a deep learning approach for Arabic spelling correction, was introduced in [44]. It used a Transformer model to learn the relationships between words and their misspellings in Arabic and was effective for Arabic spelling correction.

Recently, [45] introduced an innovative approach for Chinese spelling correction, utilizing masked language modelling to clarify the associations between correct spellings and their frequent errors. Subsequently, this method applies a correction algorithm to rectify the identified misspellings. In another work by [46], authors proposed a disentangled phonetic representation approach for Chinese spelling correction. The approach disentangled the phonetic representation of a word and used a Transformer model to learn the relationships between words and their misspellings based on the disentangled phonetic representation.

Despite the complex nature of the Persian language, the existing body of literature in Persian spelling correction is mainly comprised of statistical or rule-based methods. Mosavi and Miangah used N-grams, a monolingual corpus, and a measure of string distance to address Persian language spelling problems [47]. Kashefi and colleagues presented a new approach based on Persian keyboard character distance to prioritize correction candidates [48]. Virastyar is a Persian spelling correction software that includes several Persian NLP tools, such as a tokenizer, a POS tagger, and a spellchecker [49].

However, the spellchecker simply checked inter-word spaces and could not provide good performance. Shamsfard later reviewed Persian spelling challenges [50]. Ghayoomi and Assi created a statistical language model for the Persian language and used the model to predict the next word in Persian text typesetting to save keystrokes [51]. In another work, authors developed a hybrid method encompassing Shapex code, Soundex code, word frequencies, and a string distance metric to rank correction suggestions for the Urdu language [52]. This approach greatly increased the candidate ranking quality. However, using these hardcoded codes for each language was not cost-efficient. Vafa spellchecker is a more recent Persian spelling correction system that can detect different types of errors, including non-word, real-word, and grammar errors [53]. It leverages mutual information of word pairs based on a Persian lexicon for its scoring system and further improves its results by using N-gram probability distribution over words. Dastgheib and colleagues introduced a spelling correction system for Persian that addressed challenges specific to the language. They used the dictionary lookup method and the confusion set approach to detect non-words and real-word errors, respectively. However, their model was unable to auto-correct misspellings [54]. In a more recent work, the authors developed an automated approach for detecting and correcting misspellings in Persian radiography and ultrasound free-texts. They used a modified N-gram language model with a maximum length of four and three distinct categories of free texts [55]. While most research on Persian spelling correction focuses on non-word errors, work on detecting and auto-correcting real-word errors is still ongoing.

In reviewing the pertinent literature, several critical challenges emerge in the realm of real-word error detection and correction for Persian text. Notably, existing methods predominantly harness statistical language modeling and rule-based techniques, which, despite their foundational contributions, no longer epitomize the vanguard of research. Such conventional approaches struggle with contextual ambiguity, grapple with the limited availabilities of nuanced linguistic resources, and often inadequately address the complex morphology and syntax intrinsic to Persian. Moreover, their rigidity compromises adaptability, failing to adequately rectify the varied and evolving error patterns evident in real-world applications. Addressing these multifaceted challenges, our proposed approach advances beyond traditional paradigms, integrating a sophisticated semantic architecture with cutting-edge classifiers and a semantic ranking algorithm. This methodological advancement not only transcends the precision and contextual sensitivity attainable by the erstwhile prevalent statistical and rule-based methods but also embeds a level of adaptability and linguistic insight previously unachievable. By harnessing a dynamic analysis framework, our strategy proficiently deciphers the Persian language's intricacies, marking a significant stride over extant methods. Our meticulous evaluation across diverse corpora validates our approach's pre-eminence, delineating its robustness and efficiency in surmounting the sophisticated challenges that have long stymied advancement in the domain.

## 3. Persian Spelling Challenges

Persian, also known as Farsi, is an Indo-Iranian language of the Indo-European language family. It is the official language of Iran, Tajikistan, and Afghanistan. Persian has been influenced by other languages over time, such as Arabic, and its vocabulary has been expanded. However, the structure of the language has remained relatively unchanged for centuries [50, 56]. Persian is an expressive language, but it also presents some challenges for language processing. Here are some of the most common challenges of Persian language:

**Character ambiguity**: The Persian script, a modified version of the Arabic script, introduces character recognition ambiguities. Such ambiguities arise because certain Persian characters are visually similar, yet represent different sounds, posing challenges for accurate text processing. For example, the Persian characters ("ی" 'y' /j/) and ("ي" 'y' /j/) [3] are often used interchangeably, but they represent different sounds. This can make it difficult for language processing systems to accurately recognize and transcribe Persian text.

**Rich morphology**: Persian has a rich morphology dominated by an affixal system. This means that words can be formed by adding prefixes and suffixes to a root word. For example, the word ("کتاب" 'book' /ketɒ:b/) can be modified with the suffix ("ها" /hɒ:/) to create the plural form ("کتاب‌ها" 'books' /ketɒ:bhɒ:/). This rich morphology can make it difficult for language processing systems to accurately identify word boundaries and perform morphological analysis.

**Orthography**: Persian orthography makes use of a combination of spaces and semi-spaces (zero-width non-joiners), which are often ignored or confused, leading to orthographic inconsistencies and added sparsity. For example, the word ("می‌خورد" 'was eating' /mi:khʊrd/) is written with a semi-space between ("می" /mi:/) and ("خورد" /khʊrd/), but this space is often omitted in informal writing.

**Co-articulation**: Co-articulation, or the influence of one sound on another, can pose challenges for speech recognition and synthesis in Persian. For example, the pronunciation of the consonant ("ب" 'b' /mi:khʊrd/) can be influenced by the vowel that follows it.

**Dialectal variation**: Persian is a pluricentric language with several standard varieties,

---

[3] All pronunciations have been provided in International Phonetic Alphabet (IPA)

including Farsi (spoken in Iran), Dari (spoken in Afghanistan), and Tajik (spoken in Tajikistan) dialects. These varieties can differ significantly in terms of phonology, vocabulary, and grammar. This dialectal variation can pose challenges for language processing systems that need to accurately recognize and understand different varieties of Persian.

**Cultural factors**: The cultural context in which Persian is used can also affect language processing. The process of persianization, or the adoption of Persian language and culture by non-Persian societies, can influence the way Persian is used and understood.

**Lack of resources**: Persian is considered a low-resource language in terms of available data and tools for natural language processing. This can make it difficult to develop and evaluate language processing systems for Persian.

**Free word order**: Persian has a relatively free word order, often called scrambling. This means that words within a sentence can be rearranged without significantly changing the meaning of the sentence. For example, the sentence "من کتاب را می‌خوانم" (I am reading the book) can also be written as "کتاب را من می‌خوانم" (The book I am reading). This free word order can make it challenging for language processing systems to accurately parse sentences and identify syntactic relationships between words.

**Homophony**: Homophony occurs when different words are pronounced the same way but have different meanings. For example, ("گذاردن" /guzɒ:rdæn/ 'put') and ("گزاردن" /guzɒ:rdæn/ 'commit') are pronounced identically but have different meanings. Homophony can pose challenges for speech recognition and natural language understanding in Persian.

**Diacritics**: Diacritics are used in Persian to indicate short vowels and other phonetic features. However, they are often omitted in writing, which can lead to ambiguity in word recognition. For example, the word ("آب" 'water /ɒ:b/) can be written with or without diacritics, which can affect its meaning.

**Rapidly changing vocabulary**: Like many languages, Persian has a rapidly changing vocabulary due to the influence of technology, globalization, and other factors. This can make it difficult to keep language processing systems up-to-date with current usage.

**Lack of standardization**: There is no single standard for Persian text, which can make it difficult to develop language processing models that are able to handle a variety of different dialects and styles. For example, there are different ways to write the word "book" in Persian, such as ("کتاب" /ketɒ:b/), ("کتابی" /ketɒ:bɪ) and ("کتابچه" /ketɒ:btʃe/). This can make it difficult for language processing systems to correctly identify the word "book" in different contexts.

One of the most difficult aspects of processing Persian text is determining how to treat internal word boundaries. Internal word boundaries are usually represented by a zero-width non-joiner space, also known as a "pseudo-space." This is important because Persian words can be made up of multiple word stems, prefixes, and suffixes. If a typist does not use a pseudo-space between these word components, it can cause text processing errors. For example, the compound word "واژه‌پرداز" (pronounced "vɒ:ʒehpærdɒ:z"), which means "word processor," is often typed as "واژه پرداز" (pronounced "vɒ:ʒeh pærdɒ:z"). This is because the typist has ignored the internal word boundary between the word stems "واژه" (pronounced "vɒ:ʒeh") and "پرداز" (pronounced "pærdɒ:z"). When a pseudo-space is not used, the text processing engine will interpret these two words as separate words, which can lead to errors.

Ignoring pseudo-spaces can affect the frequency distribution of words during text processing. This is because the total frequency distribution is calculated over different word forms [56]. This problem can be resolved in a pre-processing step in which pseudo and white spaces are corrected according to internal word boundaries. Sentence tokenization is also highlighted as an open challenge in Persian language due to these problems. In addition to the word boundary challenges, tokenization problems can be addressed in the pre-processing step.

## 4. Material and Methods

In this section, we present our approach for the automatic detection and correction of real-word errors in Persian text. The logical architecture of the system consists of five distinct modules that exchange information through a databus.

The INPUT module takes the raw text as input and performs all the necessary operations, including preprocessing and normalization. The semantic module is responsible for creating the knowledge graph of a given corpus and estimating the semantic similarity between word sequences.

The real-word error detection module uses an embedded algorithm to generate various types of errors based on a desired density score. Since no known corpus of Persian word errors is available, this algorithm is used to create a synthetic corpus of errors. A classifier that relies on features from the semantic module is then used to detect real-word errors in the input text.

The error correction module uses a semantic ranking algorithm to automatically correct context-sensitive errors. This algorithm ranks the replacement candidates based on their semantic similarity to the context and their frequency in a large corpus of Persian text.

Finally, the corrected word sequence is returned through the OUTPUT module. Figure 1 depicts the logical architecture of the developed system.

The materials and methods leveraged in the proposed approach are described as follows.

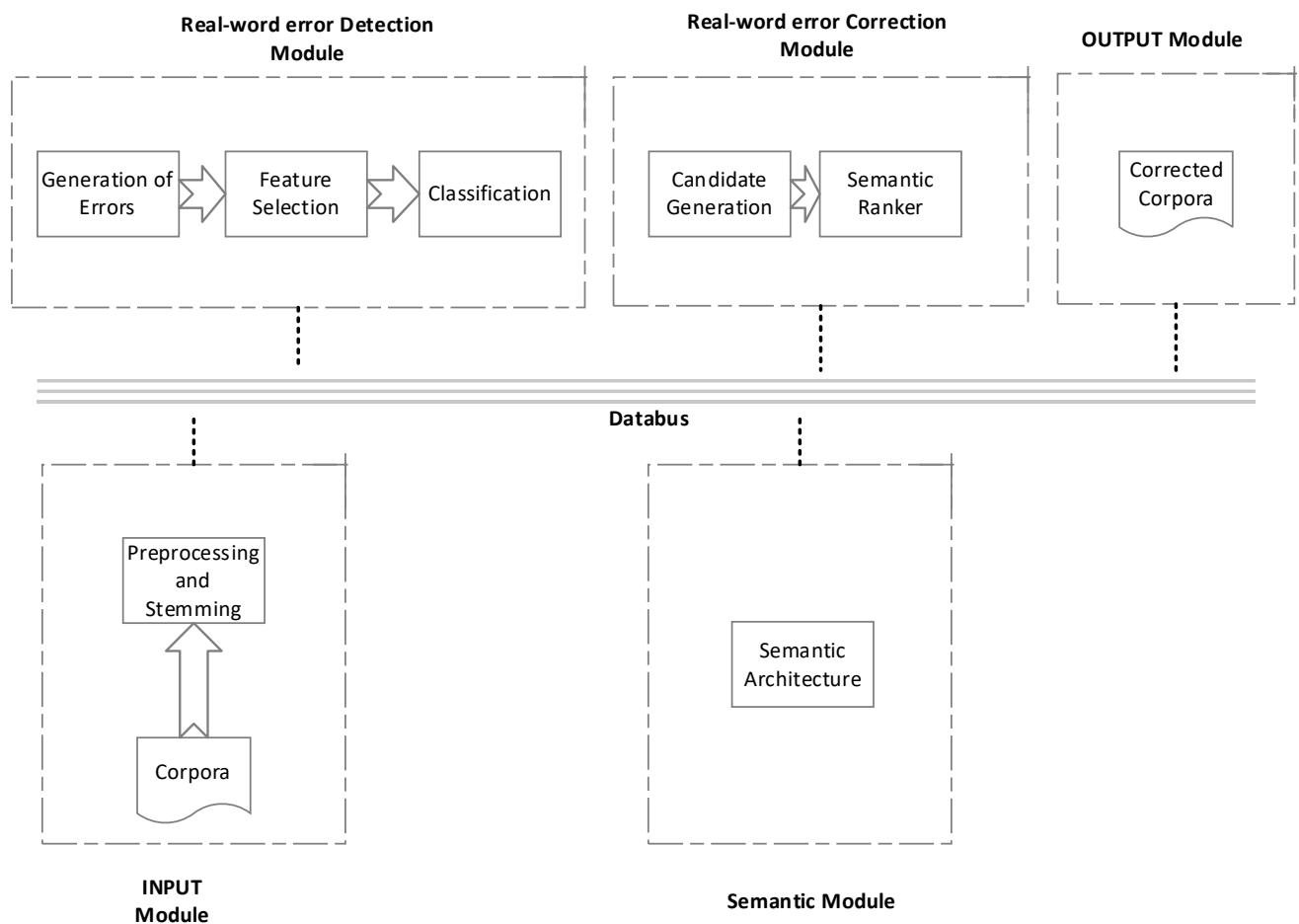

Figure 1: Logical Architecture of the Proposed System for Detecting and Correcting Real-word Errors. This diagram illustrates the system's framework, highlighting its key components and processing flow, essential for identifying and correcting real-word errors in Persian text

## 4.1 Pre-processing and Normalization Step

Text pre-processing is a crucial step in most applications of NLP. It is comprised of segmentation of sentences, tokenization, normalization, and removal of stop-words. Sentence segmentation starts with identifying the boundaries of sentences, which are generally separated by punctuation marks, namely periods, exclamation marks, or question marks. Tokenization is about splitting the sentence into a set of words that represent the sentence. This sequence of words will be used to extract the features. Normalization is the process of transforming pieces of text into canonical forms. Like many other languages, it is an essential step in NLP applications of the Persian language.

One of the important tasks in normalization of Persian text includes changing pseudo and white spaces to a standard form. Whitespaces are replaced with zero-width non-joiners whenever necessary. For example, ("می رود" /mi: rævæd/ 'goes') is replaced with ("می‌رود" /mi:rævæd/ 'goes'). As earlier mentioned, Persian and Arabic languages share similarities. There are some letters in Persian that are often misspelled by using Arabic variants. Researchers generally find it useful to make these variants normalized. For example, replacement of Arabic characters ("ی" 'Y' /j/; "ک" 'k' /k/; "ه" 'h' /h/) with their Persian equivalents; For instance ("برای" /bærɒ:ye/ 'for') is changed to ("برای" /bærɒ:ye/ 'for'). Normalization also includes removing diacritics from Persian words; e.g., ("اَره" /ærre/ 'saw') is replaced with ("اره" /ære/ 'saw'). Moreover, Kashida(s) are removed from the words; for example, ("بادبـــــادک" /bɒ:dbɒ:dæk/ 'kite') is transformed to ("بادبادک" /bɒ:dbɒ:dæk/ 'kite'). We also remove stop-words (e.g., "از" /æz/ 'from'; "به" /beh/ 'to'). These highly observed words do not carry any useful meaning. Omitting stop-words significantly improves performance.

In order to reach the normalization goal, a dictionary including the correct typographic form of all Persian words named Dehkhoda [57] is leveraged to find the normal form of multi-shaped words.

## 4.2 Stemming Step

Stemming is an essential task in many NLP applications of Persian language. This is because Persian is a morphologically rich language, meaning that words can have many different forms depending on their grammatical function. Stemming is the process of reducing a word to its stem, which is the core form of the word. For example, the words ("می‌رود" /mi:rævæd/ 'goes') and ("رفت" /ræft/ 'went') both have the stem ("رفتن" /ræftæn/ 'go').

There are three main approaches to stemming Persian words:

- Structural approaches take into account the morphological structure of the word to find the stem. For example, the structural approach proposed by [58] uses a dictionary of stems and affixes to find the stem of a word.

- Lookup table methods store all possible forms of a word in a database. The stem of a word can then be found by searching the database for the word's base form [59].

- Statistical approaches use statistical techniques to find the stem of a word. For example, the statistical approach proposed by [60] uses a machine learning algorithm to learn the probability of each affix occurring in a word.

A more recent approach to stemming Persian words is the hybrid approach proposed by [59]. This approach combines the structural and statistical approaches to achieve better performance. The hybrid approach first uses the structural approach to find a set of possible stems for a word. It then uses the statistical approach to choose the most likely stem from the set of possible stems.

The hybrid approach has been shown to outperform the other approaches to stemming Persian words. However, it is still not widely used, as it requires a large annotated corpus of Persian text to train the statistical model.

In this study, we assess the effectiveness of a hybrid stemming approach and compare its performance to that of no stemming. Our findings indicate that the hybrid method significantly enhances the performance of our spelling correction model.

### 4.3 N-gram Language Modelling and Probability Estimation

An N-gram is a statistical language modelling approach that examines sequences of *n* items in a piece of text or audio [55]. These items can be phonemes, syllables, letters, or words, depending on the application. *n* usually equals one, two, three, or four, and these are referred to as unigrams, bigrams, trigrams, and fourgrams, respectively. A simple example may be the following sequence "is it or is it not". In this example unigrams include {"is", "it", "or", "is", "it", "not"}
; the bigrams {"is it", "it or", "or is", "is it", "it not"}; and the trigrams {"is it or", "it or is", "or is it", "is it not"} and the fourgrams{"is it or is", "it or is it", "or is it not"}.

N-grams are typically generated from text or speech corpora and imply an 'intra-word' relationship. This is based on the hypothesis that word sequences that occur more frequently in the same context are considered similar. When used for language modelling, it is assumed that each word in a sentence depends only on the last $n-1$ words [61]. In other words, an N-gram language model predicts $w_i$ based on $w_{i-n+1}, \dots, w_{i-1}$. Generally N-grams are estimated based on Markov chain model as it is shown on Equation 1:

$$\prod_{i=1}^{n}(w_i | w_{i-n+1} \dots w_{i-1}) \qquad (1)$$

N-gram language models provide two main advantages: simplicity and scalability. They allow for scaling up by simply increasing the value of *n*, which provides more context about the word [62]. However, as the size of the N-gram increases, the number of unseen N-grams (sparse data) also grows, leading to a decline in the quality of the model. To address this issue, various smoothing methods have been introduced, ranging from simple Laplace smoothing to more complex models such as back-off or Good-Turing [31]. In this work, we use N-grams with a maximum length of four in the detection and correction phase.

### 4.4 Semantic similarity measures

Semantic similarity is a research area which tries to measure relatedness between word sets, concepts, sentences and documents. Similarity among sets of words is a measure of the closeness of their meaning, which is calculated based on the properties of concepts and their relationships. Similarity has an essential role in information management; particularly when data is unstructured and originate from several sources in different environments.

Measuring the semantic similarity between concepts or word sets can be leveraged in different applications including but not limited to: intelligent retrieval [63-65], word sense disambiguation [66, 67], machine learning [68], information extraction [69]. Even tough methods based on neural networks word vectors such as word2vec [70] have shown good performance in measuring semantic relatedness, yet they are behind Graph Based (knowledge or ontology-based) methods in terms of semantic similarity calculation. For instance, Qu and colleagues [71] represented that different WordNet-based methods outperform word2vec method on typical similarity. Nevertheless, how to use semantic similarity to maximize computational and increase efficiency remains an open challenge. In this work, a new semantic architecture built based on the Graph Based knowledge from Persian WordNet (or FarsNet) is used along with corpus-based probability estimation approach, for both detection and correction of context-sensitive errors. FarsNet follows the same structure as the original WordNet. [72] It is a large lexical database of Persian language, in which Nouns, verbs, adjectives and adverbs are grouped into sets of cognitive synonyms, generally known as synsets; where each express a unique concept. Moreover, a synset contains a short definition called "Gloss" where in most cases, one or more brief sentences describe the use of the synset. The first version of FarsNet included more than 10,000 synsets while the number increased to 20,000 in version 2.0and 2.5. However, the last version of FarsNet (version 3) contains more than 40,000 synsets[4].

**4.5 Design of the Semantic Architecture**

Discovering the semantic relationships between word sequences and storing them, along with probability estimates, is a complex and time-consuming process. We aim to accomplish this task using a four-level architecture:

- Data level: This level contains a pre-processed corpus, which is passed to the processor module in the processing level. Here, all N-grams of the desired order and their stemmed forms are estimated and stored in the database.
- Network level: This level contains the request sender module, which is responsible for sending the desired list of tokens to the FarsNet web service and retrieving related semantic information for these tokens.
- Information level: This level hosts the database, which includes four different tables containing various information, including unique tokens, N-grams, semantic relationships between them, and probability values.
- Processing level: The processor module resides at this level and is the heart of the system. It plays a central role in developing a semantic network by taking four main steps to create it from a training corpus:

1. The module marks sentence boundaries and identifies designated N-grams and their stemmed forms with a maximum length of four in the given word sequence.
2. It sends the desired list of tokens to the request sender module and retrieves all related semantic information for these tokens.
3. It generates a search space for all existing word sequences in the target sentence.
4. Finally, it estimates the probability of all members of the search space and stores them in the database.

Figure 2 depicts the proposed architecture; The model employs the stages in the following sections to build the semantic network. Once all elements (tokens and N-grams) of a given training corpus have been processed and the required information has been extracted, no further calls to FarsNet's Knowledge Graph are made. The system is then ready to receive test instances and process inputs while minimizing latency.

---

[4] http://farsnet.nlp.sbu.ac.ir

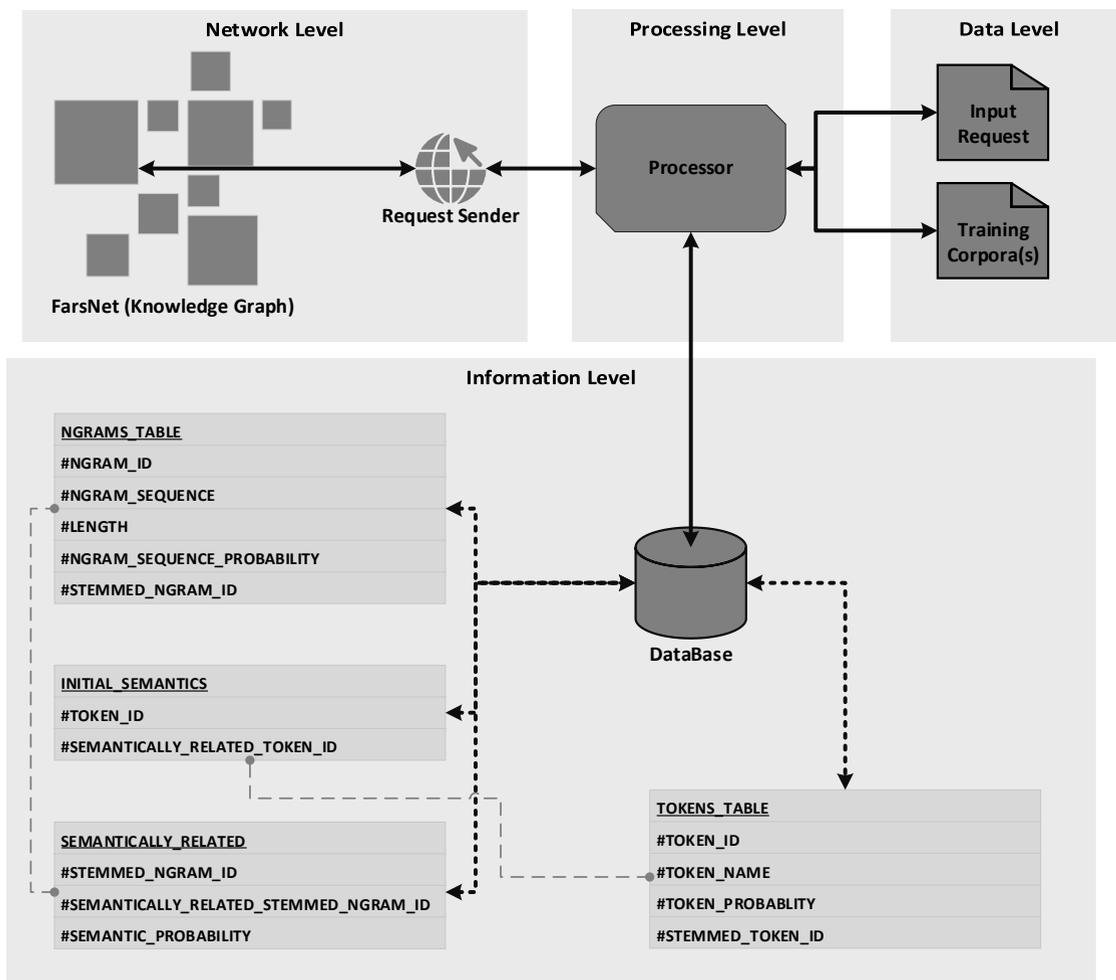

Figure 2: Four-level Architecture of the Semantic Network. This figure delineates the hierarchical structure of the semantic network, divided into four distinct levels, each playing a critical role in semantic analysis and knowledge representation. The levels are interconnected, demonstrating how information flows and is refined across the network to support advanced error detection and correction processes.

**4.5.1 Stage One: Tokenization**

At this stage, the model processes the pre-processed training corpus and saves distinct tokens and their stemmed forms, along with their probabilities, in a table named `TOKENS_TABLE`. Each token is assigned a unique ID, which serves as the primary key for the table. This table also helps to track the relationships between stemmed and inflected forms of tokens. It is important to note that the probability value for an stemmed token is equal to the sum of the probabilities of all its inflected forms in the given corpus. The tokenization and stemming stage is a crucial step in the model, as it allows the model to identify individual tokens in the corpus and calculate their probabilities. This information is then used in subsequent stages of the model to identify semantic relationships between word sequences.

**4.5.2 Stage Two: Semantic Information Retrieval**

At this stage, all related semantic information for each token is retrieved from FarsNet. FarsNet is a Persian knowledge graph that organizes words within each Part of Speech (POS) category into sets of logically related synonyms, known as synsets. Synsets are arranged in a taxonomy, ranging from the most abstract concepts to the most specific leaf concepts (end nodes). Synsets and related lexical terms are connected by pointers, which represent specific semantic relationships between them. The most commonly observed structured explicit types of relationships in FarsNet are the hyponymy (or ISA) relation and its inverse, the hypernymy relation. These two relations form the hierarchical structure of FarsNet, with the ISA relation demonstrating the generalization/specialization relationship between concepts. Another common relation used in this knowledge graph is the holonomy/meronymy (part-of) relation. Other relations, such as antonymy and derivation, are used less frequently. It is worth noting that when estimating the similarity of antonyms, they are considered to be similar, meaning that the set of words belongs to the same domain or represents features of the same concept [73]. In addition to explicit relations, lexical terms within a synset also share implicit relations, such as synonymy. This type of relation is specifically emphasized in our method.

We use the `synset_member` relation in FarsNet to identify synonyms for verbs, nouns, adjectives, and adverbs within each synset. Additionally, all antonyms, meronyms, and derivations are stored. To keep only the most semantically related information for each word, less semantically related information is excluded due to the large number of hypernyms and hyponyms. Path-based measures are used to calculate the number of edges in the shortest path between a set of concepts. The shorter the path between concepts, the more similar they are considered to be, and vice versa [67, 74, 75]. To keep the path length as short as possible, we limit it to a maximum of four edges. All related hypernyms and hyponyms within this range are retained. It is important to note that all related information is retrieved by making API calls to the FarsNet web service through the request sender module

**4.5.3 Stage Three: Semantic Information Storage**

The `INITIAL_SEMANTICS` table in the database stores the relationship between original stemmed tokens and their related semantic information, which are retrieved from FarsNet for search space generation. This table contains the original token IDs and their related token IDs, which make up the semantic sets. Each row of the table represents a semantic link between an original stemmed token and a semantically related word stem.

To avoid out-of-vocabulary words (OOVs), the model only keeps track of semantic information for tokens that have already been encountered as distinct entries in the database. This means that if a word stem is not present in the database, it will not be included in the search space. This approach helps to ensure that the search space only contains relevant and meaningful information.

### 4.5.4 Stage Four: N-gram Probability Estimation

In this step, the model runs fixed windows to extract all distinct N-grams and their stemmed forms in the given corpus. When a new sentence is received at the prompt, the model begins calculating N-gram probabilities. It uses *BoS* (Beginning of the Sentence) and *EoS* (End of the Sentence) markers to set the beginning and end of the sentence and initializes a window size parameter *d* (set to four in our case). Starting at the beginning of the sentence, a sequence of four words is accommodated by the fixed window. The existing N-grams (including unigrams, bigrams, trigrams, and fourgrams) and their stemmed versions are then designated within that window. The probabilities of N-grams and their stemmed variations are estimated using the training corpus and stored in the NGRAMS_TABLE along with the relationship between them. The window then moves one unit to the left to cover the next word sequence. If an N-gram is already present in the related database table, it is ignored and the model moves on to calculate remaining N-grams. This process is repeated until the *EoS* marker is reached. An important point to consider is that the probability of a stemmed N-gram is calculated by adding up the probabilities of all its inflected forms present in the given corpus. At the end, the NGRAMS_TABLE contains all related N-grams, their stemmed forms, and the relationship between stemmed N-grams and original N-grams, as well as the probability values for a given corpora.

### 4.5.5 Generating the Search Space and Storing the N-gram Probabilities

The search space, denoted as $C(S')$, includes all probable N-grams and their associated semantic variations based on a given sentence. The search space generation occupies a paramount position within the methodologies outlined in Sections 4.5.2, 4.5.3, and 4.5.4. The process is accomplished through the following steps:

*Step 1* – The model detects the given sentence boundaries and sets *BoS* and *EoS* markers.

*Step 2* – The value of parameter $d$ is set to four.

*Step 3* – Starting with *BoS*, a fixed window covers a sequence of four words. Related semantic information for the stemmed word sequence is retrieved from FarsNet. The relationship between the original word stem at position ***n*** in the current word sequence and its corresponding semantic information, $S_c(w_n)$, is stored in the INITIAL_SEMANTICS table. This table serves as a repository for the semantic information associated with each word in the sequence, allowing for efficient retrieval and processing of this data.
It is important to note that the model only stores instances of $S_{ci} \in S_c(w_n)$ hat can be found in the training corpora to avoid any OOV issues. As moving through hierarchies of Knowledge Graphs like FarsNet and WordNet is time-consuming, saving related data that may be frequently used will speed up the semantic set generation process. Next, all possible combinations regarding the words inside the current window are generated with respect to their order. At this step, Figure 3 and Equation 2 represent the order of instances in the search space.

$$\sum_{n=1}^{n=N} \sum_{i=0}^{i=I} W_n S_i \qquad (2)$$

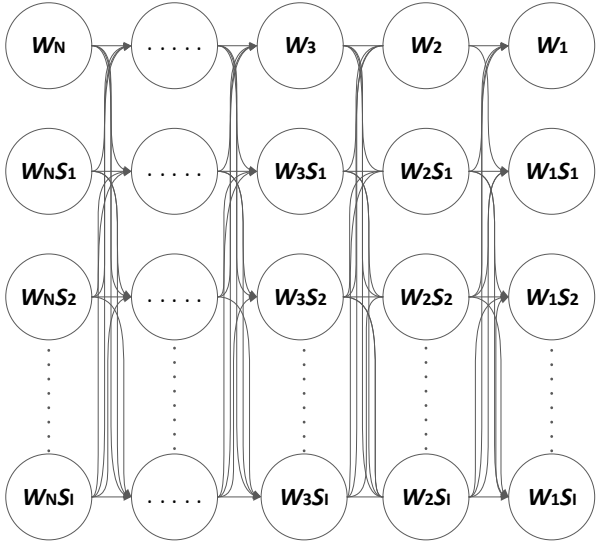

Figure 3: Order of the Search Space $C(S')$. This illustration depicts the structured hierarchy within the search space C(S'), underscoring its significance in optimizing search efficiency and system accuracy in error detection processes.

N-grams in the current window's search space $C(S')$ are a combination of a set of semantically related tokens received from FarsNet and original words, respectively. According to Figure 3, the first word in every column is the original word in the current fixed-window, with respect to the specific word order intended by the user. Other words or $S_{ci}$s n each column shape the related set of semantic information for each word $S_c(w_n)$. To be more specific, the search space for the current window, which accommodates $(w_1, w_2, w_3, w_4)$ words, includes all combinations of original words and their related semantic information with respect to their order of appearance. A combined sequence of words in the present search space is $CS_i$ with a length varying between one to four. Figure 3 demonstrates all possible combinations in the search space for a fixed-sized window, where $CS_i \in C(S')$. $W_n S_i$ is the *i*th semantically related piece of information for the *n*th word in the present window.

*Step 4* – The system extracts all N-grams within the current window, calculates their probabilities along with those of their stemmed variants, and stores this information in the NGRAMS_TABLE in the database with a unique identifier as its primary key.

*Step 5* - After discovering and estimating the probability of all the original N-grams and their regarded semantic sets in the present window; the method moves one unit to the left to covers the new words in the sentence. *Steps 4* and *5* are repeated until no remaining sentence is left in the given text.

*Step 6* - In order to establish a semantic connection between the original N-grams and their related $CS_i$s in the database, our model first verifies whether the associated $CS_i$s and their probability values are already present in the NGRAMS_TABLE. If they are, the model defines and stores the relationship between the stemmed $CS_i$ and original stemmed N-gram in the SEMANTICALLY_RELATED table. The id of the original stemmed N-gram and the id of the related stemmed $CS_i$ are stored in two adjacent columns, while the third column contains the updated probability value of the original stemmed N-gram. This value is calculated by summing up the probabilities of all related stemmed $CS_i$ in the corpora. Equation 3 represents the formula for estimating the semantic probability of a given word sequence. all words in the search space.

In our model, since all the members of a search space share the same meaning, we assign the same final semantic probability value to each of them. This value is calculated by taking the average probability value of all the word sequences in the search space. By doing so, we ensure that the final probability value accurately reflects the shared meaning of the words in the search space. As we will discuss later, this value will be used as an input feature to the classifier and semantic ranker to detect and correct real-word errors.

$$\sum_{i=1}^{i=N} \frac{\exp\left(\log\left(probability(CS_i)\right)\right)}{i} \qquad (3)$$

Utilizing this approach, our model constructs and sustains a sophisticated network that maps the intricate semantic relationships among diverse word sequences. This network forms the foundation for understanding contextual nuances and enhancing error correction accuracy.

### 4.6 Real-word Error Detection Module

The real-word error detection module employs a comprehensive multi-step process that includes error generation, feature selection, and the application of sophisticated classifiers. This structured approach ensures precise identification of real-word errors, underpinning the system's effectiveness. The following sections provide a detailed explanation of each step in our real-word error detection method, including the feature selection process and the classifiers used.

### 4.6.1 Error Generation Algorithm

Due to the scarcity of Persian texts containing authentic real-word errors, there is a significant gap in the available corpora for contextual error analysis in Persian. This scarcity hampers the development and testing of robust error detection algorithms. To address this issue, researchers have introduced artificially generated real-word errors into regular texts. For instance, in free-running English texts, the authors in [76] randomly induced malapropism into the test corpus at a frequency rate of 0.005, replacing one word in every 200 words. Different frequency rates were leveraged in [1, 27, 28]. For the Persian language, [53, 54], the authors used a method based on confusion sets; however, they did not mention the real-word error density in the test corpora.

There is also a specific constraint to the evaluation of spelling correction systems: when generating real-word errors, the maximum number of errors per sentence is set to one. As such, a sentence may contain only a single error. In our proposed model, we adopt the same approach and opt for generating a single error per sentence. Nevertheless, we experiment with different error density rates, ranging from 10% to 55% of the sentences in a test corpus containing context errors. This high density of errors was applied to measure the effectiveness of our proposed approach.

In our presented method, we use the Damerau-Levenshtein distance measure [77]. It shares the same operations - insert, delete, and substitution - with Levenshtein [10]. However, the main difference is the cost of transposition of two adjacent characters: it is 2 in Levenshtein and 1 in Damerau-Levenshtein. For instance, the Levenshtein distance $dL$(BC, CBE) equals 3, while the Damerau-Levenshtein distance $dDL$ (BC, CBE) is 2. The edit distance between two given words is the minimum number of defined operations (e.g., insertion, deletion, or substitution of a single character in case of Levenshtein distance measure) needed to change one word into another.

Damerau first reported that approximately 80% of human-generated spelling errors contained an instance of one of four error types: insertion, deletion, substitution, and transposition. [78] reported that at least 16%, according to their study, of errors within a distance of one can become a real-word error. In addition [79], reported that over half of the errors in texts are different from the correct word by an edit distance of one. This provides sufficient explanation as to why many works use a single edit distance to generate context-sensitive errors.

Although all works on real-word error correction in Persian language are based on confusion sets; even for English language we did not come across any work on context-sensitive errors that handles errors with edit distances over 1. Yet we decided to use an edit distance of up to 2 between the correct word and the context error. According to our evaluations, an average number of 5 errors are generated as replacements for a target context word when edit distance is set to 1. However, this average number grows to 23 when the distance is increased to 2. At the same time, computation time increases proportionally. Our observations show that it takes nearly eight times longer to generate errors within an edit distance of two than error words within an edit distance of one.

Pseudocode1 is leveraged to generate errors in a corpus that is presumably free of any misspellings. The algorithm generates context errors within distances of 1 and 2. The error density $E$ specifies the percentage of sentences that will contain context errors. For each target word in the text, we extract all possible instances of error that are within the desired distance in our dictionary. Our proposed model leverages a comprehensive dictionary to generate errors and replacement candidates.

The Vafa spell-checker for the Persian language dictionary is used as the main basis for this purpose. The dictionary contains over one million distinct words [53], making it a comprehensive lexicon for Persian language processing. Another parameter, *D1*, allows us to determine the proportion of errors that are within distance 1 of a correct word, with the remaining errors being of distance 2. For example, if we set $E = 0.20$ and $D1 = 0.60$, this would mean that 100 sentences in a test corpus of 500 sentences would have context errors, 60 of which would include errors with distance 1, and 40 sentences with distance 2 errors.

Pseudocode1: Generating errors in a given Persian test corpus

| | |
|---|---|
| **Input:** | Persian text corpus as $P$; Density of Error as $E$, Errors within distance 1 as $D1$; Errors within distance 2 as $1 - D1$<br>$0.1 \leq E \leq 0.55$, $0 \leq D1 \leq 1$ |
| **Output:** | Test corpus $EP$ with generated errors included in it |

| 1 | Start: | Initialize parameters $E, D1$ |
|---|---|---|
| 2 | | Recognize all sentences $s_1, s_2, ..., s_N$ boundaries |
| 3 | | Select $N*E$ sentences randomly; as random list $RL$ |
| | | *## $RL_1$ is the list of sentences where errors are within edit distance 1.* |
| 4 | | Create Random list $RL_1$; where $RL_1 \subseteq RL$ AND $RL_1 = RL * D1$ |
| | | *## $RL_2$ is the list of sentences where errors are within edit distance 2.* |
| 5 | | Create Random list $RL_2$; where $RL_2 \subseteq RL - RL_1$ AND $RL_2 = RL * (1 - D1)$ |
| | | *## Replacing target words with generated errors* |
| 6 | | Foreach $L \in RL$ do:<br>  Select word $w$ randomly from sentence $S_L$<br>  Replace $w$ with an instance $w'$ of related error set $E'$; with distance 1 if $L \in RL_1$ OR with distance 2 if $L \in RL_2$ |

### 4.6.2 Feature Selection

In our approach, we use a classifier to detect real-word errors. This classifier employs various features, including word N-

grams, stemmed N-grams (n = 1, 2, 3, and 4) which are obtained from NGRAMS_TABLE table, as well as semantic N-grams of varying lengths (1, 2, 3, and 4) retrieved from SEMANTICALLY_RELATED table. The stem-based features are dependent on a hybrid stemmer for the accurate identification and processing of word stems. As a result, we have 4 features representing the word, 4 features for the stemmed word, 4 features for stem-based related semantic sets, and finally 34 features for character frequency (to cover all written forms of Persian language characters). N-grams are widely used in many text classification tasks.

Feature selection is the process of choosing a subset of the most relevant features to be used in the classification task and model construction. It serves as a filter to discriminate among features that may not be useful in the classification task. The main goal of feature selection is to reduce computation time and improve accuracy in the classification process. There are various strategies for selecting a subset of features, including best-subset selection, forward stepwise, backward stepwise, hybrid, and embedded selection [80]. For example, best subset selection conducts an exhaustive search on all existing subsets of features to find the best model representing the best accuracy in the classification task. However, when the number exceeds 20, this approach becomes unfeasible due to the large number of feature subsets. On the other hand, forward, backward, and hybrid stepwise perform well on huge sets of feature combinations.

To evaluate feature selection algorithms, several experiments were performed using the filters Information Gain and Gain Ratio in Waikato Environment for Knowledge Analysis (Weka) [81]. In our study, we utilized three feature selection algorithms and determined that word N-grams and stem-based related semantic sets are the most prominent features. These features were ranked above character frequency and stem-based N-grams based on our results.

### 4.6.3 Classifier

The proposed model leverages support vector machines (SVMs) as a classifier for detecting real-world errors. SVMs are supervised learning models that are commonly used for binary classification, regression analysis, and outlier detection [82]. They are particularly well-suited for tasks where the number of features is much larger than the number of samples, as the complexity of the trained classifier is dependent on the number of support vectors rather than the dimensionality of the data [83]. This results in shorter classification times and better generalization performance.

In the context of real-world error detection, SVMs can be used to identify words that are misspelled or used incorrectly [84, 85]. The model learns to distinguish between correct and incorrect words by mapping them to a high-dimensional space where a linear decision boundary can be drawn between the two classes [86]. The two most important parameters that affect the performance of an SVM classifier are the penalty factor C and the kernel function. The penalty factor determines the cost of misclassifying a training sample, while the kernel function computes the similarity between two data points.

In this work, we will use a 10-fold cross-validation (CV) strategy to evaluate the performance of our SVM classifier. CV is a resampling method that is used to estimate the generalization performance of a model by training and testing it on different subsets of the data [87]. This helps to mitigate the effects of overfitting and ensures that the results are more reliable. We will experiment with both linear and non-linear SVMs, using three different kernel functions: linear, RBF (radial basis function), and sigmoid. The $\gamma$ (kernel bandwidth) hyperparameter will be set to 0.1 for the non-linear kernels. This will allow us to compare the performance of the different SVM models and select the one that performs best on our dataset.

**4.7 Real-word Error Correction Module**

One common approach to addressing real-word errors in various languages is the use of confusion sets and N-gram language models [25, 53, 54, 88, 89]. Confusion set methods work by limiting the number of errors that can be detected and corrected. However, our proposed approach does not rely on confusion sets and is capable of detecting and auto-correcting a large set of real-word errors.

In the detection phase, if the classifier identifies a word as a real-word error, the system proceeds to the correction step. In this phase, all matching correction candidates within a specific edit distance are generated. For example, if the model is set to generate errors with a distance of 1, it will attempt to use the same distance to generate a list of correction candidates. If the model is set to generate errors with a distance ≤ 2, all replacement candidates with distances of 1 and 2 are generated.

The candidates are then ranked based on their N-gram language model values (fourgram, trigram, bigram, and unigram, in that order). Finally, the semantic ranking algorithm is used to select the most probable candidate based on stem-based related semantic sets and N-gram values. Pseudocode 2 provides an explanation of the error correction process and the ranking algorithm.

اداره هواشناسی از شروع بازندگی [بارندگی] طی هفته آتی خبر داد.

The Meteorological Department has announced that there will be loss [rainfall] in the upcoming week.

In the above sentence, the word ("بازندگی" /bɒ:zændegi:/ 'loss' ) was mistakenly typed instead of the intended word ("بارندگی" /bɒ:rændegi:/ 'rainfall'). The model correctly detects this real-word error and generates a list of candidate corrections using an edit distance of 1, which was used to generate the error word in the given context. The two-stage ranker then ranks the candidates. In the first stage, the system searches for related Ngrams for each candidate, with candidates having Ngrams of the highest length being placed at the top of the list. If two or more candidates have the same length of Ngrams, the model proceeds to the second stage, where it searches for related semantic set probability values with a maximum length of 4 for each candidate. The candidates are then sorted based on their semantic set probability values, with the one having the highest value replacing the error word. In this case, ("بارندگی" /bɒ:rændegi:/ 'rain') tops the list as it frequently appears with the word ("شروع" /ʃoru:ʔ/ 'start') and is even an element of the attested word sequence " اداره هواشناسی شروع بازندگی" and its related semantic sets, more than any other correction candidates. After the correction task is accomplished, the sentence containing the error is transformed accordingly:

اداره هواشناسی از شروع بارندگی طی هفته آتی خبر داد.

The Meteorological Department has announced that there will be rainfall in the upcoming week.

Pseudocode2: The ranking algorithm for correcting the real-word errors in a given sentence.

| **Input:** | Input sentence $S$ with real-word error in it; real-word error at position $j$ of the sentence $S$, denoted as $RE_j$; edit distance $ED$; Parameters from pseudocode1 |
|---|---|
| **Output:** | Sentence $S$ with no real-word error |
| 1 Start: | Check if $RE_j \in RL_1 \; or \; if \; RE_j \in RL_2$; Initialize $ED$ based on $RE_j$; |
| 2 | Generate candidate set $C_t$ according to $ED$ where $c \in C_t$ from the dictionary; |
| 3 | Add every $C_t$ member and its stem as $c'_f$ to $C_f$<br><br>## $c'_f$ or $(C_t \cup C_t.lemmas) \subseteq C_f$ |
| 4 | Foreach $c'_f \in C_f$ do: |
| 5 | Replace the error $RE_j$ with $c'_f$, determine the transformed sentence $S'$ |
| 6 | Run the fixed-windows on the sentence $S'$ |
| 7 | Determine the target word sequences with current $c'_f$ in it |
| 8 | Find the highest related N-grams or $HRN$ and their stemmed forms to $c'_f$ from NGRAMS_TABLE table; store them with probabilities included in $HRN$ list |
| 9 | Generate the semantic sets for the present window |
| 10 | Retrieve the final probability of related semantic sets (or $CS_is$) to $c'_f$ from SEMANTICALLY_RELATED table and store it in $SHL$ |
| 11 | Sort the $c'_f$s based on the order of the N-grams in $HRN$; Keep the $c'_f$s of highest length; omit the rest of candidates |
| 12 | Foreach $c'_f \in HRN$ do: |
| 13 | Assign the semantic set probability value from $SHL$ to $c'_f$ |
| 14 | Select the $c'_f$ with largest semantic set value as the most suitable candidate to the $S'$; |

## 5. Evaluation and Results

In this section, we present a detailed evaluation of our model, including its implementation, dataset preparation, experimental results, and comparison with other baseline models. We begin by discussing the implementation of our model, including the steps taken to prepare the dataset for use in our experiments. Next, we describe the fine-tuning process for the various components of our system, highlighting the key decisions and trade-offs made during this stage. We then present the results of our evaluation, demonstrating the effectiveness of our model in achieving its intended goals and comparing its performance with other baseline models. Through this rigorous evaluation process, we are able to validate the performance of our model and provide valuable insights into its capabilities and strengths relative to other approaches.

**5.1 Model Implementation**

The system was developed using the Java programming language and utilized publicly available Persian natural language processing libraries. Text pre-processing steps were accomplished using the "normalizer" class in the jhazm library. The hybrid stemmer was also implemented in Java, and the FarsNet library was used for generating semantic sets. For machine learning tasks, including the selection of the **SVM** parameter *C* and building the **SVM** model, we used Weka, an open-source suite for data mining tasks developed in Java.

**5.2 Datasets**

Since there are no specific corpora for the spelling correction task, texts from the Hamshahri corpus version 2[5] were utilized to train and evaluate the proposed model [90]. The corpus consists of 166,774 documents divided into 82 categories and 65 subjects, with an average article length of 380 words. Although the Hamshahri corpus has 82 categories, only 12 of them contain more than 1000 articles, which cover more than 72 percent of the texts.

The majority of the documents in the Hamshahri corpus are news items from the Hamshahri newspaper that were gathered between 1996 and 2002. These articles were written by numerous authors with a variety of backgrounds and covered a wide range of topics. Additionally, this corpus includes free-running text from everyday use of the Persian language that represents appropriate sampling. The preparation procedure of related documents is thoroughly discussed in [90], where a crawler was developed to find and retrieve any online news from the Hamshahri newspaper's website. The documents in this corpus range in size from tiny news items (less than 1 KB) to lengthy articles (more than 140 KB), with an average document size of 1.8 KB.

For our evaluations, we aggregated some articles from the twelve most referenced categories - international, religious, economic, politics, social, sports, literature, science, articles, incidents, law and national security - into four distinct datasets. Table 1 represents details of each dataset.

The majority of the documents in the Hamshahri corpus are news items from the Hamshahri newspaper that were gathered between 1996 and 2002. These articles were written by numerous authors with a variety of backgrounds and covered a wide range of topics, where is a lot of information about each. Additionally, this corpus includes free-running text from everyday use of Persian language that represents the appropriate sampling. The preparation procedure of related documents is thoroughly discussed in [90], where a crawler was developed to find and retrieve any online news from the Hamshahri newspaper's website. The documents in this corpus range in size from tiny news items (less than 1 KB) to lengthy articles (more than 140 KB), with the average document size being 1.8 KB. We aggregated some articles from the twelve most referenced categories including international, religious, economic, politics, social, sports, literature, science, articles, incidents, law and national security; into four distinct datasets for our evaluations. Table 1 represents details of each dataset.

Table1: Details of dataset

---

[5] https://dbrg.ut.ac.ir/hamshahri/

| Name of the Dataset | Set1 | Set2 | Set3 | Set4 |
|---|---|---|---|---|
| # of articles | 15,712 | 6,204 | 3,455 | 69,008 |
| # of sentences | 103,840 | 86,035 | 57,638 | 123,512 |
| # of tokens | 3,496,720 | 3,191,898 | 1,700,321 | 6,546,136 |
| # of distinct tokens | 147,851 | 155,964 | 160,912 | 183,473 |

Set1 encompasses a diverse range of eight genres, including social, economic, law, national security, international, religious, sports, science, and politics, comprising a total of 103,840 sentences. Set2 covers six different news categories and includes 155,964 distinct tokens. Set3 mainly includes 3,455 articles from five different genres. Finally, Set4 is comprised of 6,546,136 non-distinct tokens that cover eleven different genres.

### 5.3 Evaluation Metrics

The main evaluation metrics used to assess the performance of models on real-word error detection and correction tasks include Accuracy (Acc), and $F-measure$ (F1). In the context of classification, these metrics are defined in terms of true positive (TP), true negative (TN), false positive (FP), and false negative (FN) comparisons of the classifier's results with prior knowledge. The labels "positive" and "negative" refer to the classifier's prediction, while "true" and "false" indicate whether this prediction matches the actual judgment.

Accuracy is the proportion of instances from both classes that were correctly classified. Precision (P) evaluates the accuracy of a classifier, while recall (R) measures its comprehensiveness or sensitivity. Precision and recall provide more detailed information about the performance characteristics of a classifier than accuracy alone. The $F-measure$ combines both metrics by calculating the weighted harmonic mean of precision and recall. In $F-measure$, precision and recall are given equal weight. Equations 4, 5, 6, and 7 describe these four evaluation metrics in detail.

$$Precision(P) = \frac{TP}{TP + FP} \quad (4)$$

$$Recall(R) = \frac{TP}{TP + FN} \quad (5)$$

$$F - measure = 2 * \frac{P * R}{P + R} \quad (6)$$

$$Accuracy(Acc) = \frac{TP + TN}{TP + TN + FP + FN} \quad (7)$$

### 5.4 Preparing the Experiments

Our performance experiments are divided into three distinct sets. For the first two sets of experiments, we will use Set1, while for the final set, we will use Set2, Set3, and Set4. The first experiment is designed to examine the effect of various system components on the performance of real-word error detection. The second set of experiments examines the effect of context error type (e.g., its edit-distance to a correct word) on correction performance. Set1 was specifically designed and utilized to accomplish these two tasks. Finally, the third set of experiments evaluates our system's performance on three different corpora: Set2, Set3, and Set4.

### 5.5 Training Corpus

The lack of a Persian language dataset with real-world errors has led to the conventional approach of injecting context error terms into sample text and then using this erroneous text to evaluate context-sensitive error detection and correction. This approach has been used by several researchers, as seen in [1, 27, 28, 53, 54].

For our training corpus, we selected a sample of 20,000 sentences from the Set1 corpus and randomly populated it with real-world errors using Pesudocode1. We chose this corpus because it is a large and comprehensive collection of Persian text.

Based on observations, most real-world errors are within one edit distance away from the intended word. This means that they can be corrected by changing one or two letters. As a result, many works have focused primarily on errors that are a single edit distance away from the correct term.

For the edit distance, we use the Damerau-Levenshtein measure, which is a more lenient measure than Levenshtein. It treats the swapping of two adjacent letters as a single operation, rather than two. This allows us to construct context errors that are within an edit-distance of 1 ($dDL = 1$) or 2 ($dDL = 2$) from a correct word.

We believe that this approach is a more realistic way to evaluate context-sensitive error detection and correction systems. It allows us to test the systems on a variety of errors, including those that are more challenging to correct.

The quantity of context errors is another issue to consider. In our experience, an adequate density for context-sensitive errors is $E = 10\%$, which means that 10% of the sentences will contain a context error while enforcing the one-error-per-sentence criterion. This amounts to a total of 2,000 incorrect words in the training corpus. However, some experiments may require a different context error density, such as $E = 25\%$ or even higher. This is achievable by adjusting the parameters of the error generation algorithm.

### 5.6 Fine-tuning System Components for Error Detection

In this section, we conduct a series of experiments to determine the impact of individual system components on context error detection performance. We investigate factors such as stemming, N-gram order, related semantic set order, edit-distance function, density of generated context errors, SVM penalty factor C, different SVM kernels, $PCA$, and stop-words.

We first set a default parameter setting for all experiments. The default settings are as follows:

1. Stemming: hybrid stemmer
2. N-gram value: $n = 1$ to 5
3. Related semantic set order: $n = 1$ to 5
4. Edit-distance function: Damerau-Levenshtein distance with $dDL = 1$
5. Density of generated context errors: 10%
6. SVM penalty factor: $1E3 \leq C \leq 1E9$
7. SVM kernel: sigmoid
8. $PCA$: no dimensionality reduction
9. Stop-words: removed

An optimal combination of stemming and SVM penalty factor C was then sought. It was observed that the hybrid stemmer outperformed the absence of stemming, and notably, the sigmoid kernel with $C = 107$ yielded the best performance.

Subsequent investigations focused on the impacts of $PCA$ and stop-words, revealing that both negatively affect performance. This outcome is attributed to $PCA$'s potential to eliminate crucial features and the non-contributory nature of frequently occurring stop-words.

Evaluation of N-gram and related semantic set order on the real-word error detection module's effectiveness was also conducted.

The linear kernel demonstrated optimal results at N-gram order 3 with an $F-measure$ of 0.889, whereas the RBF kernel reached peak performance at N-gram order 4, achieving an $F-measure$ of 0.928. Stability of the sigmoid kernel was noted across N-gram orders 3 to 5, with a maximum $F-measure$ of 0.926.

Additionally, incorporating the related semantic set property led to significant performance enhancements across all kernels. The linear kernel achieved the best performance at related semantic set order 3, with an $F-measure$ of 0.918. The RBF kernel achieved the best performance at related semantic set order 4, with an $F-measure$ of 0.955. The sigmoid kernel was relatively stable for related semantic set orders 4 to 5, with a maximum $F-measure$ of 0.965.

A comparative analysis of the results revealed that the application of the related semantic set led to a statistically significant improvement in the accuracy of real-word error detection for all three kernels. The improvement was 5.9% for the linear kernel, 2.7% for the RBF kernel, and 3.9% for the sigmoid kernel, when compared to the solo N-gram-based scheme.

We also investigated the effect of error density and edit-distance function. We observe that the performance of the linear kernel decreases as the error density increases, while the performance of the nonlinear kernels is relatively stable even when $E = 55\%$. This is likely because the non-linear kernels are better able to generalize to unseen data.

Figure 4 illustrates the impact of fine-tuning various system components on the performance of real-word error detection.

We then conducted experiments to test the ability to identify real-word errors that are more than one edit away. We created a single text with real-word errors that are either edit-distance 1 or edit-distance 2. Based on observations reported by [77], we determined that an appropriate ratio for this case would be 80% for errors that are distance 1, and 20% for errors that are distance 2.

Table 2 presents the performance of the proposed approach in terms of $F-measure$ for detecting real-word errors. The approach was evaluated using different values of the edit-distance function and error densities, based on various kernels. The results demonstrate the effectiveness of the proposed approach in accurately identifying real-word errors.

Our observations indicated that all three kernels exhibited exceptional performance when the generated context-sensitive errors were a single distance away from the original context word. However, the nonlinear kernels (RBF and sigmoid) surpassed the linear kernel in performance when the context errors were of mixed distances. The results for both kernels began to converge and improve for error densities above 30%. On the other hand, the performance of the linear kernel deteriorated further as the error density increased, due to the rise in the number of errors within distance 2.

As anticipated, the linear kernel is limited to modelling linear relationships between features, while nonlinear kernels have the ability to model more intricate relationships. This becomes particularly crucial when addressing context errors of mixed distances or distance 2, where there are a greater number of potential candidates to evaluate and manage.

Table 2: Performance of the Proposed Method in Detecting Real-Word Errors with Varying Edit-Distance Values and Error Densities on the Training Dataset

| *E* | Type of Distance | Linear kernel | RBF kernel | Sigmoid kernel |
|---|---|---|---|---|
| 10% | Edit-distance 1 | 0.948 | 0.955 | **0.965** |
| 10% | Mixed edit-distance | 0.899 | 0.952 | **0.963** |
| 20% | Edit-distance 1 | 0.925 | 0.940 | **0.943** |
| 20% | Mixed edit-distance | 0.860 | 0.935 | **0.940** |
| 30% | Edit-distance 1 | 0.908 | 0.935 | **0.939** |
| 30% | Mixed edit-distance | 0.835 | 0.928 | **0.933** |
| 40% | Edit-distance 1 | 0.897 | 0.942 | **0.948** |
| 40% | Mixed edit-distance | 0.818 | 0.937 | **0.944** |
| 50% | Edit-distance 1 | 0.880 | 0.947 | **0.960** |
| 50% | Mixed edit-distance | 0.794 | 0.942 | **0.957** |

In the experimental setup, various system components crucial for real-world error detection were investigated. Firstly, the impact of stemming was examined, with the hybrid stemmer significantly outperforming no stemming. Different SVM kernels were explored, highlighting the sigmoid kernel's superior performance. The optimal penalty factor C for the SVM was determined to be 1E7. For N-gram modelling, an N-gram order of 3 was found to be optimal for the linear kernel, while 4 was optimal for the RBF and sigmoid kernels. Similarly, the related semantic set order was determined to be 3 for the linear kernel and 4 for the RBF and sigmoid kernels. Regarding error density, non-linear kernels exhibited greater robustness to high error densities, particularly at an error density of 10%. Lastly, concerning the edit-distance function, nonlinear kernels (RBF and sigmoid) showed superior performance, especially in scenarios involving mixed distance errors, utilizing the Damerau-Levenshtein distance with $dDL = (1,2)$. These findings underscore the importance of selecting appropriate hyperparameters tailored to the characteristics of the data and task at hand in error detection system.

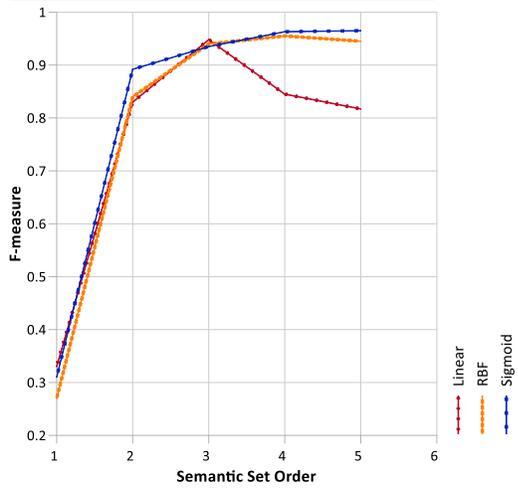
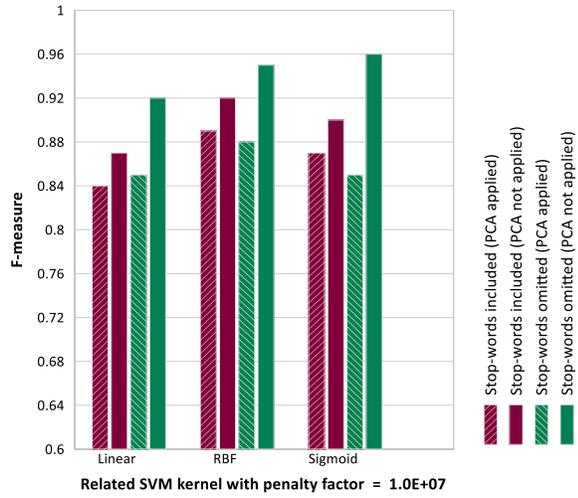
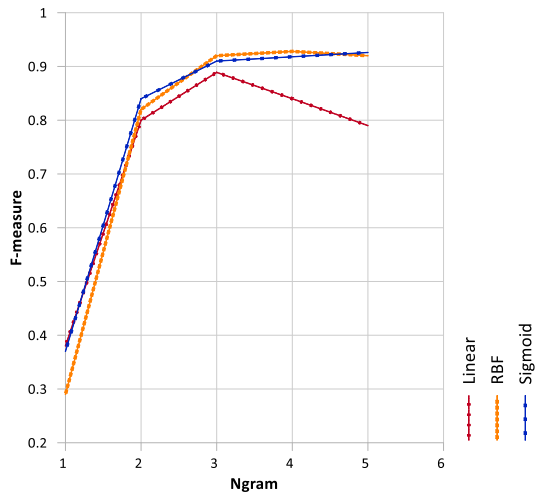
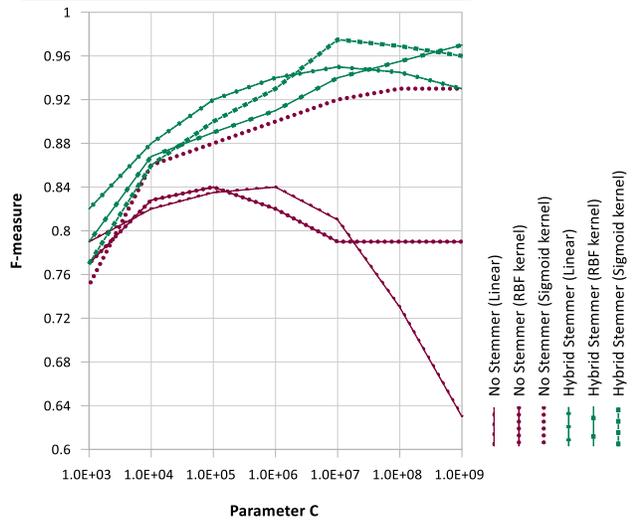
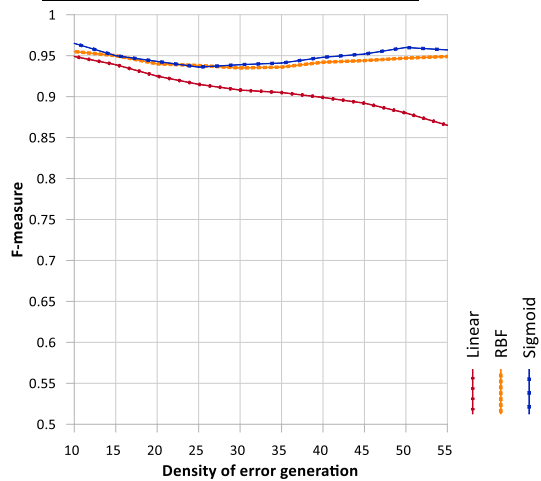

Figure 4: Impact of Fine-tuning Various System Components on the Performance of Real-word Error Detection. This chart illustrates the enhancements in real-word error detection efficiency achieved through meticulous fine-tuning of system components, emphasizing their vital roles in the system's overall effectiveness.

## 5.7 Fine-tuning Impact on Error Correction

The next step is to replace the erroneous word with the most likely candidate. To evaluate the effectiveness of the semantic ranking algorithm, we employ the F-measure formula, as detailed in Equation 6. In this evaluation, we set up two scenarios: (1) finding the candidate with the highest order N-grams and largest probability values, and (2) leveraging related semantic set probability values.

Table 3 presents the percentage of context error words in a sample corpus of 20,000 sentences, with varying error density values, that were accurately corrected using the default settings described in section 4.6. Two different groups of context errors were used in our experiments: (1) contextual errors with an edit distance of one ($dDL = 1$) from a correct word, and (2) those with a mix of one and two (80% with one and 20% with two).

The real-word error correction module demonstrates remarkable performance when the edit distance between the context word and the generated matching words is confined to 1, and the error density is set at 10%. In this scenario, the $F - measure$ value is 0.975 when related semantic sets are not utilized and 0.991 when they are implemented. However, there is a significant decline in the $F - measure$ value when the edit distance encompasses both 1 and 2, particularly when the correction module relies solely on estimating the original N-grams in the given sentence. In this situation, 93.1% and 83.8% of incorrect words within distances 1 and 2 are successfully rectified, respectively. The overall correction $F - measure$ for mixed errors, considering distances 1 and 2, indicates that 91.24% of detected context errors have been successfully amended. Conversely, the accuracy of correcting mixed-distance errors with the application of the semantic property is noteworthy, with an $F - measure$ of 0.953. This represents a 4.1% increase compared to the no-semantics scheme.

We conducted an exploration into the impact of higher error densities on the $F - measure$ of error correction models. We compared two distinct error densities: a minimum error density of 10% and a maximum error density of 50%.

Our results, presented in table 3, reveal that when the semantic set property is not applied, the model is more susceptible to errors in selecting the correct replacement candidate, particularly at higher error densities. For instance, the difference between $F - measure$ values for selecting the proper candidate within distance 1 for minimum and maximum values of $E$ is 0.053. This indicates a decrease in accuracy of 5.3% when the error density escalates from 10% to 50%.

In contrast, when the semantic set property is applied, our model exhibits increased robustness to higher error densities. The difference in $F - measure$ values for selecting the proper candidate within distance 1 for minimum and maximum values of $E$ reduces to 0.025, indicating only a decrease in accuracy of 2.5% when the error density escalates from 10% to 50%.

The performance of our model is also promising when rectifying detected errors of mixed distances. For example, without implementing the semantic set property, the difference between $F - measure$ values for selecting the proper candidates of mixed distance for minimum and maximum error densities is 0.043, indicating a decrease in accuracy of 4.3% when the error density increases from 10% to 50%. However, with the implementation of the semantic set property, this difference reduces to 0.020, indicating only a decrease in accuracy of 2%.

The property of semantic sets equips the model with the ability to analyze sequences of words based on their inherent semantic meaning, rather than solely on their orthographic similarity. The empirical evidence from this study highlights that the integration of semantically related sets into our real-word error correction algorithm can significantly enhance its performance. This improvement is evident even when the algorithm confronts complex errors that are challenging to rectify. By utilizing this supplementary information, our model achieves superior accuracy and robustness in correcting real-world errors.

Table 3: Performance of the Proposed Method in Correcting Real-Word Errors with Varying Edit-Distance Values on the Training Dataset

| *E* | Semantic Set | Edit-Distance 1 | Mixed-distance | | |
|---|---|---|---|---|---|
| | | | edit-distance1 | edit-distance2 | overall result |
| 10% | *not applied* | **0.975** | 0.931 | 0.838 | 0.912 |
| 10% | *applied* | **0.991** | 0.961 | 0.924 | 0.954 |
| 20% | *not applied* | **0.961** | 0.919 | 0.825 | 0.900 |
| 20% | *applied* | **0.981** | 0.952 | 0.916 | 0.945 |
| 30% | *not applied* | **0.938** | 0.900 | 0.807 | 0.881 |
| 30% | *applied* | **0.973** | 0.946 | 0.910 | 0.939 |
| 40% | *not applied* | **0.927** | 0.893 | 0.798 | 0.874 |
| 40% | *applied* | **0.968** | 0.943 | 0.907 | 0.936 |
| 50% | *not applied* | **0.922** | 0.888 | 0.794 | 0.869 |
| 50% | *applied* | **0.966** | 0.941 | 0.906 | 0.934 |

## 5.8 Baseline Models

In this section, we replicate the methodology of prominent real-word error correction models for the Persian language. It is important to note that all the baseline models were implemented using the Java programming language.

We implemented two baseline models for comparison purposes. Firstly, we replicated and coded the real-word error correction models based on confusion sets, as proposed by Dastgheib et al. [54]. Secondly, we implemented the real-word error correction module in the Vafa spell-checker [53].

### 5.8.1 Vafa Spell-checker

The Vafa spell-checker is designed to detect and correct real-word errors in Persian text using a three-step process:
1. Context analysis: The spell-checker initiates its process by examining the context of the words, taking into account the words that come before and after it.
2. Candidate Generation: Next, the spell-checker generates a list of candidate words for the misspelled word by considering all possible

single-letter edits and semantically similar words.
3. Confusion Set: This approach relies on predefined confusion sets to detect real-word errors. The model replaces word sequence tokens with members of the same confusion set and examines the resulting N-gram in the training corpora.

We implemented this model using a trigram language model.

### 5.8.2 Perspell

Perspell is a statistical spelling correction pipeline for the Persian language that utilizes a dictionary look-up strategy to detect non-word errors and a bigram language model to identify the most suitable candidate in a given context. For real-word error detection, this model relies on the use of predefined confusion sets and synonym retrieval from Persian WordNet. It then employs a simple bigram language model to examine probable word sequences and correct the error.

However, Perspell is primarily designed for candidate suggestion rather than error correction. To make the model suitable for auto-correction, we made some amendments. After sorting the candidates based on their probability value, the most probable word was selected as the replacement candidate.

### 5.9 Models Evaluation on full datasets

In this section, we evaluate the detection and correction capabilities of our system in comparison to the developed baselines. We selected three distinct corpora, namely Set2, Set3, and Set4, which are significant datasets comprising approximately 86,000, 58,000, and 124,000 sentences, respectively. We then artificially introduced real-word errors into these sentences using two different error densities: 10% and 50% ($E = 10\%, 50\%$). The first error density simulates a typical source of error (e.g., a typist), while the second error density replicates an extremely noisy source. We ensured that all generated errors were of mixed distances, with 80% at distance 1 and the remaining at distance 2 ($dDL = 1,2$).

For our model evaluation, we used the following setup: a hybrid stemmer was employed, $PCA$ was not utilized, stop words were eliminated, the N-gram language model had an order of n = 1 to 4, the semantic set order ranged between 1 and 4, 10-fold cross-validation was applied, the SVM penalty factor was $C = 1E7$, and the SVM kernel was sigmoid.

Table 4 presents the results of evaluating our approach and the baselines' detection performance on various corpora. We observe that our proposed approach achieved its best performance on the Set3 dataset, which contains the fewest number of test instances for both error density values, with an $F - measure$ of 0.966. The method proved to be robust as the difference in overall $F - measure$ for evaluation on the largest dataset with $E = 50\%$ and the smallest dataset with $E = 10\%$ is only 1.6%. The method also demonstrated promising performance in detecting errors at both distances 1 and 2.

The authors of [50] claimed to have achieved an $F - measure$ of 0.726 for real-word error correction; however, we were unable to replicate this result in our evaluations. We challenged all models against a large number of test instances to evaluate their practical performance. In the case of Perspell, the maximum $F - measure$ of 0.644 was obtained for detecting real-word errors in the Set3 dataset with $E = 10\%$. These results indicate that the model is not very effective in detecting errors at distance 2. In our evaluation, Vafa Spell-checker achieved the least favorable results with a maximum $F - measure$ of 0.154 on the Set3 dataset. Our results indicate that it was not effective in detecting real-word errors at both distances 1 and 2.

In addition to the results presented in Table 4, we conducted a qualitative analysis of the errors made by our model. We found that the model was more likely to misdetect errors when the artificially generated error was semantically related to the context words. For instance, in the original word sequence "برای انتقال اثر" (to transfer the effect), the error generation algorithm replaced the word ("اثر" /æsær/ 'effect') with the artificially generated error ("ارز" /ærz/ 'currency'), which is within edit distance 1. This resulted in the word sequence " برای انتقال ارز" (to transfer currency), which had a higher N-gram and semantic probability value than the original word sequence. Consequently, the model ignored this word sequence.

Although this issue has not been reported in previous Persian spelling correction research, we believe it is a significant problem when populating Persian corpora with artificially generated errors. This is likely due to the lack of genuine error corpora in Persian. However, this issue could be mitigated by checking generated errors against the list of N-grams and semantic probability values in the error generation algorithm.

In summary, our proposed method performed well on all datasets; however, there were some minor differences in performance related to the size of the dataset. The method achieved state-of-the-art results in real-word error detection. It is robust to different error densities and dataset sizes and performs well at both distances 1 and 2.

Table 4: Evaluation Results of Proposed Model and Baselines for Real-Word Error Detection Task in Terms of F-measure on Three Distinct Corpora

| E | Model | Corpora | Mixed-distance | | |
|---|---|---|---|---|---|
| | | | edit-distance1 | edit-distance2 | overall result |
| 10% | Proposed method | Set2 | 0.971 | 0.935 | **0.964** |
| 10% | Perspell | Set2 | 0.636 | 0.549 | 0.619 |
| 10% | Vafa Spell-checker | Set2 | 0.153 | 0.126 | 0.148 |
| 50% | Proposed method | Set2 | 0.965 | 0.929 | **0.958** |
| 50% | Perspell | Set2 | 0.587 | 0.514 | 0.572 |
| 50% | Vafa Spell-checker | Set2 | 0.131 | 0.102 | 0.125 |
| 10% | Proposed method | Set3 | 0.973 | 0.938 | **0.966** |
| 10% | Perspell | Set3 | 0.663 | 0.570 | 0.644 |
| 10% | Vafa Spell-checker | Set3 | 0.161 | 0.128 | 0.154 |
| 50% | Proposed method | Set3 | 0.966 | 0.932 | **0.959** |
| 50% | Perspell | Set3 | 0.631 | 0.537 | 0.612 |
| 50% | Vafa Spell-checker | Set3 | 0.149 | 0.124 | 0.144 |
| 10% | Proposed method | Set4 | 0.963 | 0.928 | **0.956** |
| 10% | *Perspell* | **Set4** | 0.606 | 0.536 | 0.592 |
| 10% | *Vafa Spell-checker* | **Set4** | 0.141 | 0.107 | 0.137 |
| 50% | *Proposed method* | **Set4** | 0.956 | 0.924 | **0.950** |
| 50% | *Perspell* | **Set4** | 0.561 | 0.475 | 0.544 |
| 50% | *Vafa Spell-checker* | **Set4** | 0.124 | 0.100 | 0.119 |

We also evaluated the ability of all models to correct real-word errors. The results in table 5 demonstrate that our proposed method outperforms the other models in correcting real-word errors at various distances and densities. The highest $F-measure$ of 0.960 is achieved on the Set3 dataset at a 10% error density, while the lowest $F-measure$ of 0.929 is obtained on the Set4 dataset with an error density of 50%. The difference of 3.1% is reasonable given the large number of corrections made. The results are also satisfactory for correcting context-errors at distance 2, suggesting that our proposed method is robust and precise in correcting detected context-errors.

The Perspell correction software achieves the best F-measure of 0.321 for correcting real-word errors on the Set3 dataset, with an error density of 10%. However, its performance declines significantly when applied to the Set4 dataset with a 50% error density, resulting in a decrease of 5.1% in $F-measure$. This suggests that Perspell frequently fails to replace the detected word errors with the correct candidate. The results are even worse for correcting real-word errors at distance 2.

The Vafa Spell-checker shows the poorest performance among the three models, with a maximum $F-measure$ of 0.287 achieved on the Set3 corpora with an error density of 10%. Its performance is even more disappointing in correcting context-errors at distance 2.

We have identified some cases where our system failed to identify the correct replacement word. Our semantic ranker simply picks the top word in the ranked correction candidates list. However, this may not always be the correct word that the user intended.

For example, consider the sentence "علاوه بر منهای قانون است" (addition to the minus the law). The system correctly identified the context error word, ("منهای" /menhɒ:y/ 'minus'). The correction is ("مبنای" /mæbnɒ:ye/ 'basis'), but the ranked candidates list is: ("مبنای" /mæbnɒ:ye/ 'basis'), ("معنای" /mænɒ:ye/ 'definition) and ("مینای" /mi:nɒ:ye/ 'enemal'). The intended correction is the second entry in the list, but the N-gram and semantic probability values of the first and second entries are close. This means that either of the top two words in the ranked candidates list could be the correct choice.

To address this issue, we could give characters weight based on keyboard similarity or similar phonetics. For example, the characters "ع", "ن" and "ه" are very close on the keyboard, so we could give them more weight when comparing the two words "معنای" and "منهای".

Overall, our proposed approach demonstrates superior performance in correcting context-errors, while the two baseline models are deficient in accuracy and practicality for application.

Table 5: F-measure Results of Real-Word Error Correction Task for Proposed Model and Baselines on Three Datasets

| E | Model | Corpora | Mixed-distance | | |
|---|---|---|---|---|---|
| | | | edit-distance1 | edit-distance2 | overall result |
| 10% | *Proposed method* | **Set2** | 0.964 | 0.926 | **0.956** |
| 10% | *Perspell* | **Set2** | 0.318 | 0.289 | 0.312 |
| 10% | *Vafa Spell-checker* | **Set2** | 0.287 | 0.264 | 0.282 |
| 50% | *Proposed method* | **Set2** | 0.944 | 0.908 | **0.937** |
| 50% | *Perspell* | **Set2** | 0.279 | 0.252 | 0.274 |
| 50% | *Vafa Spell-checker* | **Set2** | 0.255 | 0.230 | 0.250 |
| 10% | *Proposed method* | **Set3** | 0.968 | 0.929 | **0.960** |
| 10% | *Perspell* | **Set3** | 0.327 | 0.295 | 0.321 |
| 10% | *Vafa Spell-checker* | **Set3** | 0.292 | 0.268 | 0.287 |
| 50% | *Proposed method* | **Set3** | 0.947 | 0.910 | **0.940** |
| 50% | *Perspell* | **Set3** | 0.288 | 0.259 | 0.282 |
| 50% | *Vafa Spell-checker* | **Set3** | 0.262 | 0.235 | 0.257 |
| 10% | *Proposed method* | **Set4** | 0.957 | 0.922 | **0.950** |
| 10% | *Perspell* | **Set4** | 0.307 | 0.278 | 0.301 |
| 10% | *Vafa Spell-checker* | **Set4** | 0.277 | 0.256 | 0.273 |
| 50% | *Proposed method* | **Set4** | 0.936 | 0.903 | **0.929** |
| 50% | *Perspell* | **Set4** | 0.275 | 0.251 | 0.270 |
| 50% | *Vafa Spell-checker* | **Set4** | 0.247 | 0.223 | 0.242 |

## 6. Discussion

Our investigation elucidates a comprehensive approach for real-word error correction in Persian, underpinned by a semantic framework with classifiers and a semantic ranking algorithm, aiming to improve the precision of both error detection and correction processes. We conducted evaluations across four distinct corpora, each characterized by error densities that ranged from 10% to 50%. The results from these evaluations provide evidence of the model's effectiveness, demonstrating error detection and correction rates that are notably better than those reported in existing research, with F-measure scores of 96.6% and 99.1%, respectively. Furthermore, the methodology has shown to be adaptable to variations in error densities and dataset sizes, maintaining efficient performance for edit distances of 1 and 2.

This approach has several potential applications, including use in spell checkers, text editors, and machine translation systems. It could also be used to enhance the quality of Persian text on the internet and in other digital media.

Our proposed method has several advantages over previous methods. Firstly, it is built upon a flexible semantic architecture that allows the model to process, store, and integrate information from multiple corpora of Persian text, making it more adaptable to new data. Secondly, the approach takes advantage of semantic similarity measures and considers the semantic context of word sequences, enabling the model to better handle real-word errors. Results demonstrate that application of the semantic set property leads to a 3.9% and 4.1% increase in detection F-measure (when sigmoid kernel is employed) and correction F-measure respectively. Thirdly, the model is capable of handling different types of real-word errors, including substitution, deletion, transposition, and insertion errors.

We conducted a qualitative analysis of the errors made by our model and found that it was more likely to misdetect errors when the artificially generated error was semantically related to the context words. In these cases, the error had a higher N-gram and semantic probability value than the original word, causing the model to ignore such word sequences. We believe that this is likely due to the lack of genuine error corpora in Persian. To address this limitation, we could check the generated errors against the list of N-grams and semantic probability values in the error generation algorithm.

Another limitation of our study is the selection of the best candidate for correction. We found that our system sometimes fails to identify the correct replacement word. Our semantic ranker simply chooses the top word in the ranked list of correction candidates, but this may not always be the word that the user intended. This issue can be addressed by using specific spelling correction properties, such as keyboard similarity or similar phonetics.

In addition to these limitations, our model is only designed to handle real-word errors and cannot handle other types of errors, such as non-word and grammatical errors.

Despite these limitations, our proposed approach represents a significant advancement in the field of real-word error detection and correction in Persian text. It outperforms baseline models and is resilient to varying error densities and dataset sizes. While there is room for improvement, our approach provides a promising foundation for developing a more accurate and robust real-world error correction system for Persian text.

## 7. Conclusion

In this study, we presented a novel methodology for the detection and correction of real-word errors in Persian text, employing an advanced semantic architecture that has demonstrated superior efficacy over existing approaches. The method achieved notable F-measures of 96.6% and 99.1% for detection and correction, respectively, and showcased robustness against variations in error density and dataset sizes. This approach significantly contributes to the enhancement of Persian text processing and comprehension.

Future work will focus on investigating advanced machine learning techniques, such as recurrent neural networks, and augmenting our model with additional features like phonetic and keyboard similarity to address a broader array of errors, including non-word and grammatical inaccuracies. These initiatives are aimed at further refining the accuracy and utility of our methodology, thereby extending its relevance and application in the field of computational linguistics for Persian language resources.

## Data Availability

The dataset supporting this article are from previously reported studies and datasets (Hamshahri corpus), which have been cited. The data are available at:

https://dbrg.ut.ac.ir/hamshahri/

## Conflicts of Interests

The author(s) declare(s) that there is no conflict of interest regarding the publication of this paper

## Funding Statement

The research did not receive any specific funding

## References

1. Wilcox-O'Hearn, A., G. Hirst, and A. Budanitsky. *Real-word spelling correction with trigrams: A reconsideration of the Mays, Damerau, and Mercer model*. in *International conference on intelligent text processing and computational linguistics*. 2008. Springer.
2. Hirst, G. and A. Budanitsky, *Correcting real-word spelling errors by restoring lexical cohesion.* Natural Language Engineering, 2005. **11**(1): p. 87-111.
3. Deng, L. and X. Huang, *Challenges in adopting speech recognition.* Communications of the ACM, 2004. **47**(1): p. 69-75.
4. Jurafsky, D. and H. James, *martin, J. 2008. Speech and Language Processing: An Introduction to Natural Language Processing.* Computational Linguistics, and Speech Recognition, 2nd ed. New Jersey: Prentice-Hall.
5. Bassil, Y. and M. Alwani, *Ocr context-sensitive error correction based on google web 1t 5-gram data set.* arXiv preprint arXiv:1204.0188, 2012.
6. Hartley, R.T. and K. Crumpton, *Quality of OCR for degraded text images.* arXiv preprint cs/9902009, 1999.
7. Huang, Y., Y.L. Murphey, and Y. Ge. *Automotive diagnosis typo correction using domain knowledge and machine learning*. in *2013 IEEE Symposium on Computational Intelligence and Data Mining (CIDM)*. 2013. IEEE.
8. Kukich, K. *Techniques for automatically correcting words in text(abstract)*. in *ACM Annual Computer Science Conference: Proceedings of the 1993 ACM conference on Computer science*. 1993.
9. FJ., D., *A technique for computer detection and correction of spelling errors.* Communications of the ACM, 1964 **3**(7): p. 171-176.
10. Levenshtein, V.I. *Binary codes capable of correcting deletions, insertions, and reversals*. in *Soviet physics doklady*. 1966. Soviet Union.
11. Atkinson, K., *Gnu aspell 0.60. 4*. 2006, GNU Aspell) Retrieved from http://aspell. net.
12. Idzelis, M. and B. Galbraith, *Jazzy: The java open source spell checker.* 2005, Retrieved 2019/10/10, from http://jazzy. sourceforge. net.
13. Crowell, J., et al., *A frequency-based technique to improve the spelling suggestion rank in medical queries.* Journal of the American Medical Informatics Association, 2004. **11**(3): p. 179-185.
14. Mitton, R., *Ordering the suggestions of a spellchecker without using context.* Natural Language Engineering, 2009. **15**(2): p. 173-192.
15. Turchin, A., et al. *Identification of misspelled words without a comprehensive dictionary using prevalence analysis*. in *AMIA Annual Symposium Proceedings*. 2007. American Medical Informatics Association.
16. Church, K.W. and W.A. Gale, *Probability scoring for spelling correction.* Statistics and Computing, 1991. **1**(2): p. 93-103.
17. Flor, M. and Y. Futagi. *On using context for automatic correction of non-word misspellings in student essays*. in *Proceedings of the seventh workshop on building educational applications Using NLP*. 2012.
18. Lai, K.H., et al., *Automated misspelling detection and correction in clinical free-text records.* Journal of biomedical informatics, 2015. **55**: p. 188-195.
19. Norvig, P., *Natural language corpus data.* Beautiful data, 2009: p. 219-242.
20. Wilbur, W.J., W. Kim, and N. Xie, *Spelling correction in the PubMed search engine.* Information retrieval, 2006. **9**(5): p. 543-564.
21. Dashti, S.M.S., et al., *Toward a Thesis in Automatic Context-Sensitive Spelling Correction.* 2014.
22. Cauteruccio, F., et al., *Extraction and analysis of text patterns from NSFW adult content in Reddit.* Data & Knowledge Engineering, 2022. **138**: p. 101979.
23. Bonifazi, G., et al., *A Space-Time Framework for Sentiment Scope Analysis in Social Media.* Big Data and Cognitive Computing, 2022. **6**(4): p. 130.
24. Mays, E., F.J. Damerau, and R.L. Mercer, *Context based spelling correction.* Information Processing & Management, 1991. **27**(5): p. 517-522.
25. Samanta, P. and B.B. Chaudhuri. *A simple real-word error detection and correction using local word bigram and trigram*. in *Proceedings of the 25th conference on computational linguistics and speech processing (ROCLING 2013)*. 2013.
26. Wilcox-O'Hearn, L.A., *Detection is the central problem in real-word spelling correction.* arXiv preprint arXiv:1408.3153, 2014.
27. Dashti, S.M., A. Khatibi Bardsiri, and V. Khatibi Bardsiri, *Correcting real-word spelling errors: A new hybrid approach.* Digital Scholarship in the Humanities, 2018. **33**(3): p. 488-499.
28. Dashti, S.M., *Real-word error correction with trigrams: correcting multiple errors in a sentence.* Language Resources and Evaluation, 2018. **52**(2): p. 485-502.
29. Kilicoglu, H., et al. *An ensemble method for spelling correction in consumer health questions*. in *AMIA Annual Symposium Proceedings*. 2015. American Medical Informatics Association.
30. Pande, H. *Effective search space reduction for spell correction using character neural embeddings*. in *Proceedings of the 15th Conference of the European Chapter of the Association for Computational Linguistics: Volume 2, Short Papers*. 2017.